\documentclass{article}

\usepackage{algorithm,algpseudocode} %algorithm
\usepackage{amsmath}
\usepackage{enumerate}
\usepackage{verbatim}
\usepackage{bm}
\usepackage{array}
\usepackage{hyperref} %urls, hyperlinks
\usepackage{xspace}
\usepackage{color}
\usepackage{graphicx} %include figures
\usepackage{caption}
\usepackage{cleveref}
\usepackage{enumerate}
\usepackage{subcaption}
\usepackage{amssymb} %mathbb

\usepackage[utf8]{inputenc}

\DeclareRobustCommand{\ie}{\textit{i}.\textit{e}.\@\xspace}
\DeclareRobustCommand{\eg}{\textit{e}.\textit{g}.,\@\xspace}

\newcommand{\xiao}{\textcolor{black}}

\newcolumntype{P}[1]{>{\centering\arraybackslash}p{#1}}

\begin{document}
\title{Active Learning for Visual Question Answering:\\ An Empirical Study}
\date{}
%\author{Xiao Lin \\Virginia Tech \\ {\tt linxiao@vt.edu} and Devi Parikh \\ Georgia Tech \\parikh@gatech.edu}

\author{
\begin{tabular}{P{5cm} P{5cm}}
\large Xiao Lin & \large Devi Parikh \\
Virginia Tech & Georgia Tech \\
{\tt linxiao@vt.edu} & {\tt parikh@gatech.edu}
\end{tabular}
}

\maketitle

\begin{abstract}

We present an empirical study of active learning for Visual Question Answering, where a deep VQA model selects informative question-image pairs from a pool and queries an oracle for answers to maximally improve its performance under a limited query budget. Drawing analogies from human learning, we explore cramming (entropy), curiosity-driven (expected model change), and goal-driven (expected error reduction) active learning approaches, and propose a new goal-driven active learning scoring function to pick question-image pairs for deep VQA models under the Bayesian Neural Network framework. We find that deep VQA models need large amounts of training data before they can start asking informative questions. But once they do, all three approaches outperform the random selection baseline and achieve significant query savings. For the scenario where the model is allowed to ask generic questions about images but is evaluated only on specific questions (\eg questions whose answer is either yes or no), our proposed goal-driven scoring function performs the best.

\end{abstract}

%%%%%%%%% BODY TEXT
\section{Introduction}
\label{sec:al_intro}

Visual Question Answering (VQA) ~\cite{Antol_2015,Gao_2015,Geman_2015,Goyal_2016,Malinowski_2015,Ren_2015} is the task of taking in an image and a free-form natural language question and automatically answering the question.
Correctly answering VQA questions arguably demonstrates machines' image understanding, language understanding and and perhaps even some commonsense reasoning abilities. 
Previous works have demonstrated that deep models which combine image, question and answer representations, and are trained on large corpora of VQA data are effective at the VQA task.

\begin{figure}[t]
\centering
   \caption{Performance of two representative VQA models: LSTM+CNN~\cite{Lu_2015} and HieCoAtt~\cite{Lu_2016} on random subsets of the VQA v1.0 dataset. Both models improve by 12\% with every order of magnitude of more training data.}
   \includegraphics[width=0.65\linewidth]{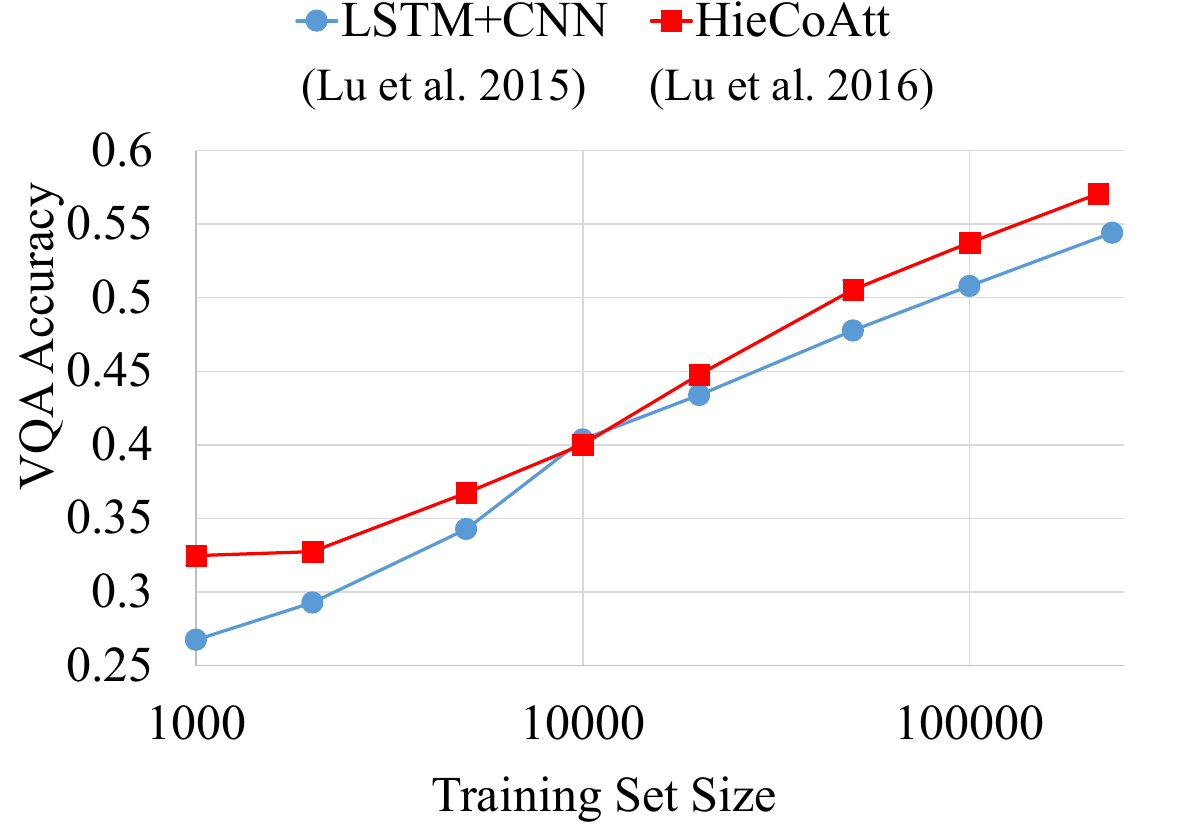}
\label{fig:fig1}
\end{figure}

Although such deep models are often deemed data-hungry, the flip-side is that their performance scales well with more training data. In Fig.\ref{fig:fig1} we plot performance versus training set size of two representative deep VQA models: LSTM+CNN~\cite{Lu_2015} and HieCoAtt~\cite{Lu_2016} trained on random subsets of the VQA v1.0 dataset~\cite{Antol_2015}. We see that for both methods, accuracy improves significantly -- by 12\% -- with every order of magnitude of more training data. 
As performance improvements still seem linear, it is reasonable to expect another 12\% increase by collecting a VQA dataset 10 times larger. Such trends are invariant to choice of image feature~\cite{Kafle_2017} and is also observed in ImageNet image classification~\cite{Mishkin_2016}. Note that improvements brought by additional training data may be orthogonal to improvements in model architecture. 

However, collecting large quantities of annotated data is expensive. Even worse, as a result of long tail distributions, it will likely result in redundant questions and answers while still having insufficient training data for rare concepts. This is especially important for learning commonsense knowledge, as it is well known that humans tend to talk about unusual circumstances more often than commonsense knowledge which can be boring to talk about~\cite{Gordon_2013}. Active learning helps address these issues. In active learning, a model is first trained on an initial training set. It then iteratively expands its training set by selecting potentially informative examples according to a query strategy, and seeking annotations on these examples. Previous works have shown that a carefully designed query strategy effectively reduces annotation effort required in a variety of tasks for shallow models. For deep models however, active learning literature is scarce and mainly focuses on classical unimodal tasks such as image and text classification.

In this work we study active learning for deep VQA models. VQA poses several unique challenges and opportunities for active learning. 

First, VQA is a multimodal problem. Deep VQA models may combine Multi-Layer Perceptrons (MLPs), Convolutional Neural Nets (CNNs), Recurrent Neural Nets (RNNs) and even attention mechanisms to solve VQA. Such models are much more complex than MLPs or CNNs alone studied in existing active learning literature and need tailored query strategies.

Second, VQA questions are free-form and open-ended. In fact, VQA can play several roles from answering any generic question about an image, to answering only specific question types (\eg questions with ``yes/no'' answers, or counting questions), to being a submodule in some other task (\eg image captioning as in \cite{Lin_2016}). Each of these different scenarios may require a different active learning approach.

%Although technically one can ask any question about images and learn anything related to the image, but in practice VQA can be used for many different purposes such as answering generic or specific questions or even used as a submodule to improve performance on other tasks. Accuracy on VQA questions per se may not be the end goal. Each of these different scenarios may require different active learning approaches.
Finally, VQA can be thought of as a Visual Turing Test~\cite{Geman_2015} for computer vision systems. To answer questions such as ``does this person have 20/20 vision'' and ``will the cat be able to jump onto the shelf'', the computer not only needs to understand the surface meaning of the image and the question, but it also needs to have sufficient commonsense knowledge about our world. \xiao{One could argue that proposing informative questions about images is also a test of commonsense knowledge and intelligence.}

We draw coarse analogies to human learning and explore three types of information-theoretic active learning query strategies: 

\textbf{Cramming -- maximizing information gain in the training domain.} The objective of this strategy is to efficiently memorize knowledge in an unlabeled pool of examples. This strategy selects unlabeled examples whose label the model is most uncertain about (maximum entropy).

\textbf{Curiosity-driven learning -- maximizing information gain in model space.} The objective of this strategy is to select examples that could potentially change the belief on the model's parameters  (also known as expected model change). There might exist examples in the pool whose labels have high uncertainty but the model does not have enough capacity to capture them. In curiosity-driven learning the model will skip these examples. BALD~\cite{Gal_2017,Houlsby_2011} is one such strategy for deep models, where examples are selected to maximize the reduction in entropy over model parameter space. 

\textbf{Goal-driven learning -- maximizing information gain in the target domain.} The objective of this strategy is to gather knowledge to better achieve a particular goal (also known as expected error reduction). To give an example from image classification, if the goal is to recognize digits i.e., the target domain is digit classification, dog images in the unlabeled pool are not relevant even though their labels might be uncertain or might change model parameters significantly. On the other hand, in addition to digit labels, some other non-digit labels such as the orientation of the image might be useful to the digit classification task. We propose a novel goal-driven query strategy that computes mutual information between pool questions and test questions under the Bayesian Neural Network~\cite{Blundell_2015,Gal_2015} framework. 

We evaluate active learning performance on VQA v1.0~\cite{Antol_2015} and v2.0~\cite{Goyal_2016} under the pool-based active learning setting described in Section
\ref{sec:al_approach}. We show that active learning for deep VQA models requires a large amount of initial training data before they can achieve better scaling than random selection. In other words, the model needs to have enough knowledge to ask informative questions.
But once it does, all three querying strategies outperform the random selection baseline, saving 27.3\% and 19.0\% answer annotation effort for VQA v1.0 and v2.0 respectively. Moreover, when the target task is restricted to answering only ``yes/no'' questions, our proposed goal-driven query strategy beats random selection and achieves the best performance out of the three active query strategies.

\section{Related Work}
\subsection{Active Learning}

Active learning query strategies for shallow models~\cite{Settles_2010,Krishnakumar_2007} often rely on specific model simplifications and closed-form solutions. Deep neural networks however, are inherently complex non-linear functions. This poses challenges on uncertainty estimation.

In the context of deep active learning for language or image understanding, ~\cite{Zhou_2010} develops a margin-based query strategy on Restricted Boltzmann Machines for review sentiment classification. ~\cite{Krause_2016} queries high-confidence web images for active fine-grained image classification.  \cite{Sener_2017} proposes a query strategy based on feature space covering, applied to deep image classification. 
Closest to our work, ~\cite{Gal_2017} studies BALD~\cite{Houlsby_2011}, an expected model change query strategy computed under the Bayesian Neural Network~\cite{Blundell_2015,Gal_2015} framework applied to image classification.

In this work we study active learning for VQA. VQA is a challenging multimodal problem. Today's state-of-the-art VQA models are deep neural networks. We take an information-theoretic perspective and study three active learning objectives: minimizing entropy in training domain (entropy), model space (expected model change) or target domain (expected error reduction). \xiao{Drawing coarse analogy from human learning, we call them cramming, curiosity-driven and goal-driven learning respectively. We apply the Bayesian Neural Network~\cite{Blundell_2015,Gal_2015} framework to compute these strategies. In particular, for goal-driven learning which was deemed impractical beyond binary classification on small datasets~\cite{Settles_2010}, we propose a fast and effective query scoring function that speeds up computation by hundreds of millions of times, and show that it is effective for VQA which has $1,000$ classes and $>400,000$ examples on contemporary multi-modal deep neural nets. } 

\subsection{Visual Conversations}

Building machines that demonstrate curiosity -- machines that improve themselves through conversations with humans -- is an important problem in AI.

~\cite{Mostafazadeh_2016,Mostafazadeh_2017} study generating human-like questions given an image and the context of a conversation about that image.
~\cite{Strub_2017} uses reinforcement learning to learn an agent that plays a ``Guess What?'' game~\cite{De_2016}: finding out which object in the image the user is looking at by asking questions. 
~\cite{Das_2017} studies grounded visual dialog~\cite{Das_2016} between two machines in collaborative image retrieval, where one machine as the ``answerer'' has an image and answers questions about the image while the other as ``questioner'' asks questions to retrieve the image at the end of the conversation. Both machines are learnt to better perform the task using reinforcement learning.

In this work we study visual ``conversations'' from an active learning perspective. In each round of the conversation, a VQA model strategically chooses an informative question about an image and queries an oracle to get an answer.  Each round of ``conversation'' provides a new VQA training example which improves the VQA model.

\section{Approach}
\label{sec:al_approach}

We study a pool-based active learning setting for VQA: A VQA model is first trained on an initial training set $\mathcal{D}_{train}$. 
It then iteratively grows $\mathcal{D}_{train}$ by greedily selecting batches of high-scoring question-image pairs $(Q,I)$ from a human-curated pool according to a query scoring function $s(Q,I)$.
The selected $(Q,I)$ pairs are sent to an oracle for one of $J$ ground truth answers $A\in \{a_1,a_2,\ldots,a_J\}$, and $(Q,I,A)$ tuples are added as new examples to $\mathcal{D}_{train}$. 
\footnote{VQA models require a large training set to be effective. To avoid prohibitive data collection cost and focus on evaluating active learning query strategies, in this work we study pool-based active learning which makes use of existing VQA datasets. Having the model select or even \emph{generate} questions for images it would liked answered, as opposed to picking from a pool of $(Q,I)$ pairs is a direction for future research.} 

We take an information-theoretic perspective and explore cramming, curiosity-driven, and goal-driven query strategies as described in Section~\ref{sec:al_intro}. However computing $s(Q,I)$ for those query strategies directly is intractable, as they require taking expectations under the model parameter distribution. So in Section~\ref{sec:al_bayesian_vqa} we first introduce a Bayesian VQA model which enables variational approximation of the model parameter distribution. And then Section~\ref{sec:strategy} introduces the query scoring functions and their approximations.

\subsection{Bayesian LSTM+CNN for VQA}
\label{sec:al_bayesian_vqa}
We start with the LSTM+CNN VQA model~\cite{Lu_2015}. The model encodes an image into a feature vector using the VGG-net~\cite{Simonyan_2014} CNN, encodes a question into a feature vector by learning a Long Short Term Memory (LSTM) RNN, and then learns a multi-layer perceptron on top that combines the image feature and the question feature to predict a probabilistic distribution over top $J=1000$ most common answers. 

In order to learn a variational approximation of the posterior model distribution, we adopt the Bayesian Neural Network framework~\cite{Blundell_2015,Gal_2015} and introduce a Bayesian LSTM+CNN model for VQA. Let $\bm{\omega}$ be the parameters of the LSTM and the multi-layer perceptron (we use a frozen pre-trained CNN). We learn a weight-generating model with parameter $\bm{\theta}$:

\begin{align}
\nonumber \bm{\omega} & =\bm{\theta} \circ \bm{\epsilon} \\
\epsilon_i & \sim \textnormal{Bernoulli}(0.5)
\end{align}

Let $q_\theta(\bm{\omega})$ be the probabilistic distribution of weights generated by this model. Following~\cite{Blundell_2015,Gal_2015}, we learn $\bm{\theta}$ by minimizing KL divergence $KL(q_\theta(\bm{\omega})||p(\bm{\omega}|\mathcal{D}_{train}))$ so $q_\theta(\bm{\omega})$ serves as a variational approximation to the true model parameter posterior $p(\bm{\omega}|\mathcal{D}_{train})$. Specifically we minimize

\begin{align}
\label{eq:al_kl}
 KL(q_\theta(\bm{\omega})||p(\bm{\omega}|\mathcal{D}_{train})) = \underbrace{\mathbb{E}_{\bm{\omega} \sim q_\theta(\bm{\omega})} \lbrack -\log P(\mathcal{D}_{train}|\bm{\omega}) \rbrack }_{\text{Cross entropy loss}} + \underbrace{KL(q_\theta(\bm{\omega})||p(\bm{\omega}))}_{\text{Deviation from weight prior}} 
\end{align}

using batch Stochastic Gradient Descent (SGD) to learn $\bm{\theta}$. In practice, $KL(q_\theta(\bm{\omega})||p(\bm{\omega}))$ can be naively approxmiated with a parametric hybrid $L1$ - $L2$ norm~\cite{Gal_2015}. Experiments show that such an naive approximation does not have a significant impact on active learning results. So in experiments we set this term to 0. How to come up with a more informative prior term is an open problem for Bayesian Neural Networks.

Let $P(A|Q,I,\bm{\omega})$ be the predicted $J$-dimensional answer distribution of the VQA model for question-image pair $(Q,I)$ when using $\bm{\omega}$ as model parameters. A Bayesian VQA prediction for $(Q,I)$ using variational approximation $q_\theta(\bm{\omega})$ is therefore given by: 

\begin{align}
P(A=a|Q,I) \approx \mathbb{E}_{\bm{\omega}\sim q_\theta(\bm{\omega})} P(A=a|Q,I,\bm{\omega})
\end{align}

\subsection{Query Strategies and Approximations}
\label{sec:strategy}

We experiment with 3 active learning query strategies: cramming, curiosity-driven learning and goal-driven learning.

\textbf{Cramming} \xiao{or ``uncertainty sampling''~\cite{Settles_2010}} minimizes uncertainty (entropy) of answers for questions in the pool. It selects $(Q,I)$ whose answer $A$'s distribution has maximum entropy. This is a classical active learning approach commonly used in practice.  

\begin{align}
\label{eq:al_s_entropy}
\nonumber s_{entropy}(Q,I) & = \mathbb{H}(A) \\
& = -\sum_a P(A=a|Q,I) \log P(A=a|Q,I)
\end{align}

\textbf{Curiosity-driven learning} or ``expected model change'' minimizes uncertainty (entropy) of model parameter distribution $p(\bm{\omega}|\mathcal{D}_{train})$. It selects $(Q,I)$ whose answer $A$ would expectedly bring steepest decrease in model parameter entropy if added to the training set.
\begin{align}
\label{eq:al_s_curiosity}
\nonumber s_{curiosity}(Q,I) & =  \mathbb{H}(\bm{\omega})-\mathbb{H}(\bm{\omega}|A)  \\
\nonumber &= \mathbb{I}(\bm{\omega};A) \\
&=\mathbb{H}(A)-\mathbb{H}(A|\bm{\omega})
\end{align}

Intuitively, $\mathbb{H}(A)-\mathbb{H}(A|\bm{\omega})$ computes the divergence of answer predictions under different model parameters. If plausible models are making divergent predictions on a question-image pair $(Q,I)$, the answer to this $(Q,I)$ would rule out many of those models and thereby reduce confusion.

According to BALD~\cite{Gal_2017}, the conditional entropy term $\mathbb{H}(A|\bm{\omega})$ in Eq.~\ref{eq:al_s_curiosity} can be approximated by:
\begin{align}
\mathbb{H}(A|\bm{\omega}) \approx -\mathbb{E}_{\bm{\omega}\sim q_\theta(\bm{\omega})} \sum_a P(A=a|Q,I,\bm{\omega}) \log P(A=a|Q,I,\bm{\omega})
\end{align}

\textbf{Goal-driven learning} or ``expected error reduction'' minimizes uncertainty (entropy) on answers $A'_t$ to a given set of unlabeled test question-image pairs $(Q'_t, I'_t), t=1,2,...T$, against which the model will be evaluated. The goal-driven query strategy selects the pool question-image pair $(Q,I)$ that has the maximum total mutual information with $(Q'_t,I'_t), t=1,2,...T$. That is, it queries $(Q,I)$ pairs which maximize:

\begin{align}
\label{eq:al_s_goal_old}
\nonumber s_{goal}(Q,I) = & \sum_{t} \mathbb{H}(A'_t) - \mathbb{H}(A'_t|A) \\
\nonumber = & \sum_{t} \mathbb{I}(A;A'_t) \\
= &\sum_{t} \sum_a \sum_{a'} P(A=a,A'_t=a'|Q,I,Q'_t,I'_t) \log \frac{P(A=a,A'_t=a'|Q,I,Q'_t,I'_t)}{P(A=a|Q,I)P(A'_t=a'|Q'_t,I'_t)}
\end{align}

For term $P(A=a,A'_t=a'|Q,I,Q'_t,I'_t)$, observe that when the model parameter $\bm{\omega}$ is given, $(Q,I)$ and $(Q'_t,I'_t)$ are two different VQA questions so their answers -- $A$ and $A'_t$ respectively -- are predicted independently. In other words, $A$ and $A'_t$ are independent conditioned on $\bm{\omega}$. Therefore we can take expectation over model parameter $\bm{\omega}$ to compute this joint probability term:

\begin{align}
\label{eq:joint_prob_approx}
\nonumber & P(A=a,A'_t=a'|Q,I,Q'_t,I'_t) \\
\nonumber & = \mathbb{E}_{\bm{\omega}} P(A=a|Q,I,\bm{\omega})P(A'_t=a'|Q'_t,I'_t,\bm{\omega}) \\
& \approx \mathbb{E}_{\bm{\omega}\sim q_\theta(\bm{\omega})} P(A=a|Q,I,\bm{\omega})P(A'_t=a'|Q'_t,I'_t,\bm{\omega})
\end{align}

\xiao{Let $M$ be the number of samples of $\bm{\omega}$, $J$ be the number of possible answers, and $U$ be the number of examples in the pool. Computing $\mathbb{I}(A;A'_t)$ for all $U$ examples in the pool following Eq.~\ref{eq:joint_prob_approx} has a time complexity of $O(UTJ^2M)$. For VQA typically the pool and test corpora each contains hundreds of thousands of examples and there are 1000 possible answers, \eg $U=400\text{k}$, $T=100\text{k}$ and $J=1,000$. We typically use $M=50$ samples in our experiments. So computing Eq.~\ref{eq:joint_prob_approx} is still time-consuming and can be prohibitive for even larger VQA datasets.
To speed up computation, we approximate $\text{log}(\cdot)$ using first-order Taylor expansion and discover that the following approximation holds empirically (more details can be found in Appendix~\ref{appendix:al_goal_approx} and~\ref{appendix:al_quality}):  }

\begin{algorithm}[t]
  \caption{Active learning for Visual Question Answering}
 \label{alg:al}
  \begin{algorithmic}[1]
  \State Initialize $\mathcal{D}_{train}$ with $N$ inital training examples. Use the rest of $(Q,I)$ in VQA TRAIN set as pool.Q
  \State Train $\bm{\theta}$ on $\mathcal{D}_{train}$ for $K$ epochs using Eq.~\ref{eq:al_kl} for initial $q_{\bm{\theta}}(\bm{\omega})$.
  \For{$iter=1,\ldots,L$}
	\State Sample $\bm{\omega} \sim q_{\bm{\theta}}(\bm{\omega})$ $M$ times.
	\State Using each $\bm{\omega}$ to make predictions $P(A|Q,I,\bm{\omega})$ on all pool and test question-image pairs.
	\State Compute $s(Q,I)$ for every $(Q,I)$ in pool using Eq.~\ref{eq:al_s_entropy},~\ref{eq:al_s_curiosity} or~\ref{eq:al_s_goal}.
	\State Select the top $G$ high-scoring $(Q,I)$ pairs from the pool.\footnotemark
	\State Lookup answers $A$ for $(Q,I)$ pairs in the VQA training set (proxy for querying a human).
	\State Add $(Q,I,A)$ tuples to $\mathcal{D}_{train}$.
	\State Update $\bm{\theta}$ on new $\mathcal{D}_{train}$ for $K$ epochs. \label{step10}
  \EndFor
  \end{algorithmic}
\end{algorithm}

\begin{align}
\label{eq:al_s_goal}
\nonumber  s_{goal}&(Q,I) \\
\nonumber \approx & \frac{1}{2} \Big[  \mathbb{E}_{\bm{\omega}} \mathbb{E}_{\bm{\omega}'} \sum_a \frac{P(A=a|Q,I,\bm{\omega}) P(A=a|Q,I,\bm{\omega}')}{P(A=a|Q,I)} \\
 & \sum_t \sum_a \frac{P(A'_t=a|Q'_t,I'_t,\bm{\omega}) P(A'_t=a|Q'_t,I'_t,\bm{\omega}')}{P(A'_t=a|Q'_t,I'_t)} -T \Big]
\end{align}

\xiao{Eq.~\ref{eq:al_s_goal} brings drastic improvements to time complexity. It can be computed as a dot-product between two vectors of length $M^2$. One only involves pool questions $(Q,I)$. The other one only involves test questions $(Q_t',I_t')$ and can be precomputed for all pool questions. Precomputing vectors for test questions has a time complexity of $O(TJM^2)$. Computing Eq.~\ref{eq:al_s_goal} using the precomputed test vector has a time complexity of $O(UJM^2)$. So the overall time complexity is linear to dataset size $max(U,T)$ and the number of possible answers $J$.}

\footnotetext{{Jointly selecting a batch of $(Q,I)$ pairs that optimizes the active learning objectives may further improve active learning performance. Deriving query strategies that can select batches of examples under the Bayesian Neural Network framework is part of future work. }}

In previous works, goal-driven learning was deemed impractical beyond binary classification on small datasets~\cite{Settles_2010}. Previous works explore the goal-driven learning objective for shallow classifiers such as Naive Bayes~\cite{Roy_2001}, Support Vector Machines~\cite{Guo_2007} and Gaussian Process~\cite{Zhu_2003}. However on VQA, computing such scoring functions would require learning a new set of model parameters for every possible combinations of $(Q,I,A)$ and then making predictions on all $(Q'_t, I'_t)$ using the learnt model. That would require $4 \times 10^{15}$ forward-backward passes (10 billion epochs) for VQA neural nets.  Instead our Monte-Carlo approximation of Eq.~\ref{eq:al_s_goal} only involves making predictions on $(Q,I)$ and $(Q'_t,I'_t)$, and avoids training new models for each of $J=1,000$ answers when computing $s_{goal}(Q,I)$. In our approach, the operation with the highest time complexity is a matrix multiplication operation which in practice is not the bottleneck. The most time-consuming operation -- computing scores for $P(A|Q,I,\bm{\omega})$ and $P(A'_t|Q'_t,I'_t,\bm{\omega})$ -- costs approximately $3 \times 10^{7}$ forward passes (75 epochs), a speed up of more than $10^8$ times. Our approach is easily parallelizable and works for all Bayesian Neural Networks.

%\begin{align}
%\mathbb{I}(A;A'_t) \approx \frac{1}{2} 
%\end{align}

Our active learning procedure is summarized in Algorithm~\ref{alg:al}.

\section{Experiment}
\label{sec:al_exp}
\subsection{Experiment Setup}

\begin{figure*}[t]
\centering
   \caption{Active learning versus passive learning on (top) VQA v1.0 and (bottom) v2.0. All three active learning strategies perform better than passive learning. Best viewed in color.}
   \includegraphics[width=1\linewidth]{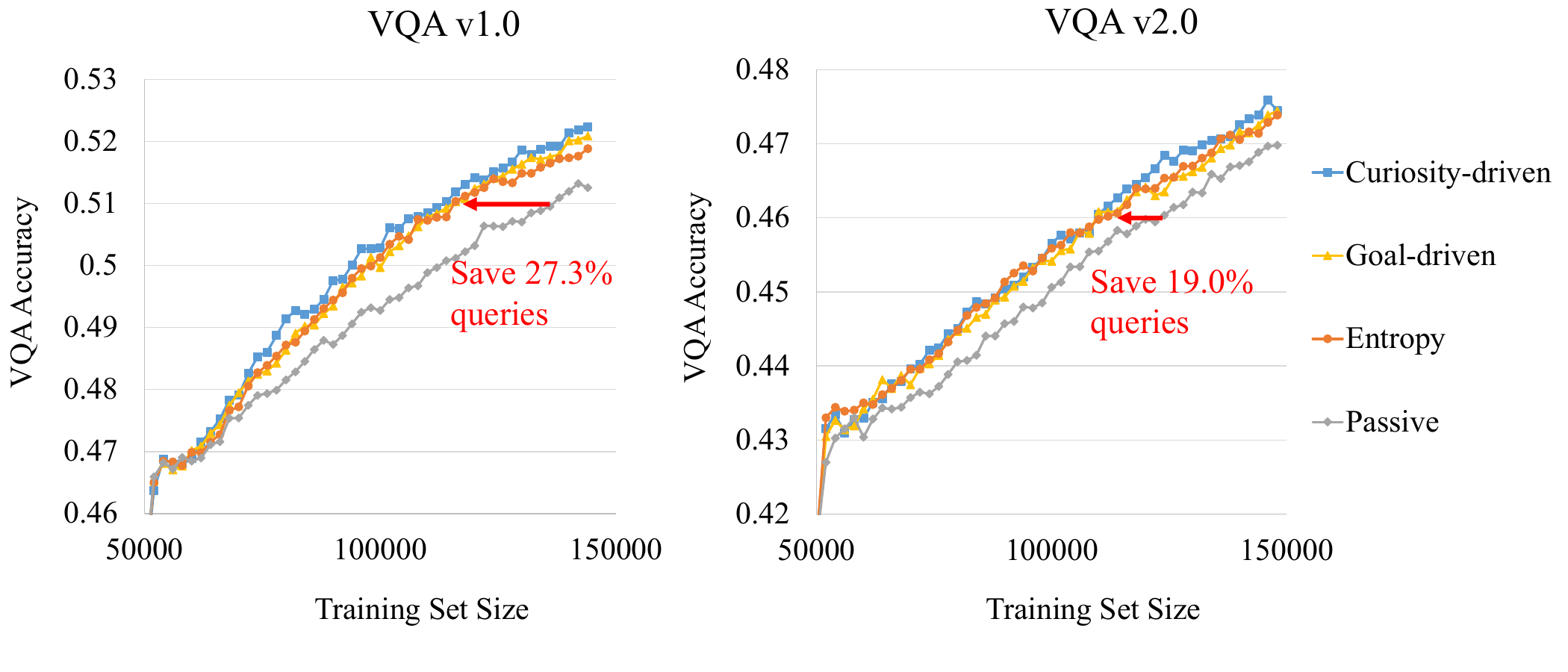}
\label{fig:al_main}
\end{figure*} 

We evaluate cramming (entropy), curiosity-driven and goal-driven active learning strategies against passive learning on the VQA v1.0~\cite{Antol_2015} and v2.0~\cite{Goyal_2016} datasets. The VQA v1.0 dataset consists of 614,163 VQA questions with human answers on 204,721 COCO~\cite{TYLin_2014} images. The VQA v2.0 dataset augments the VQA v1.0 dataset and brings dataset balancing: every question in VQA v2.0 is paired with two similar images that have different answers to the question. So VQA v2.0 doubles the amount of data and models need to focus on the image to do well on VQA v2.0.

We choose a random initial training set of $N=50\text{k}$ $(Q,I)$ pairs from the TRAIN split, use the rest of TRAIN as pool and report VQA accuracy~\cite{Antol_2015} on the VAL split. We run the active learning loop for $L=50$ iterations. We sample model parameter $\bm{\omega}$ for $M=50$ times for query score computation. For passive learning \ie querying $(Q,I)$ pairs randomly, we set $s_{passive}(Q,I)\sim \text{uniform}(0,1)$. In each iteration $G=2,000$ $(Q,I,A)$ pairs are added to $\mathcal{D}_{train}$, resulting in a training set of $150\text{k}$ examples by the end of iteration 50.

For VQA model, we use the Bayesian LSTM+CNN model described in Section~\ref{sec:al_bayesian_vqa}. In every active learning iteration we train the model for $K=50$ epochs with learning rate $3\times10^{-4}$ and batch size $8\times128$ for learning $q_{\bm{\theta}}(\bm{\omega})$.

\subsection{Active Learning on VQA v1.0 and v2.0}

Fig.~\ref{fig:al_main} (left), (right) show the active learning results on VQA v1.0 and v2.0 respectively. On both datasets, all 3 active learning methods perform similarly and all of them outperform passive learning. On VQA v1.0, passive learning queries 88k answers before reaching 51\% accuracy, where as all active learning methods need only 64k queries, achieving a saving of 27.3\%. It shows that active learning is able to effectively tell informative VQA questions from redundant questions, even among high-quality questions generated by humans. Similarly at 46\% accuracy, active learning on VQA v2.0 achieves a saving of 19.0\%. Savings on VQA v2.0 is lower, possibly because dataset balancing in VQA v2.0 improves the informativeness of even a random example.

\begin{table}[t]
\centering
\caption{On VQA v2.0 for each pair of query strategy, what percentage of $(Q,I)$ pairs are selected by both methods. Active learning (entropy, curiosity-driven, goal-driven) query strategies select $>80\%$ common $(Q,I)$ pairs and they are very different from passive learning.}
\begin{tabular}{ l | c c c c }
\hline
$(Q,I)$    & Passive  & Entropy & Curiosity & Goal \\
Overlap (\%)   &  learning &  & driven & driven \\
\hline
Passive learning &  - & 26.70  & 26.65  & 26.64 \\
Entropy  & 26.70 & -  & 83.26 & 82.52 \\
Curiosity-driven & 26.65 & 83.26 & - & 85.27 \\
Goal-driven & 26.64 & 82.52 & 85.27 & - \\
\hline
\end{tabular}
\label{table:overlap}
\end{table}

\xiao{Table~\ref{table:overlap} shows that for each pair of active learning methods, what percentage of the query $(Q,I)$ pairs are selected by both methods on VQA v2.0 (overlap between their training sets). For the VQA task, active learning methods seem to agree on which $(Q,I)$ pairs are more informative. They have more than 80\% of $(Q,I)$ pairs in common, while against passive learning they only share $\sim$27 \% of $(Q,I)$ pairs. }

\begin{figure}[htbp]
\centering
   \caption{Active learning with $N=20\text{k}, 10\text{k}, 5\text{k}, 2\text{k}$ initial training set size. When dataset size is small, active learning is unable to outperform passive learning. The breakpoint when active learning methods start to perform better is around $30\text{k}$ to $50\text{k}$ examples. Best viewed in color.}
   \includegraphics[width=1\linewidth]{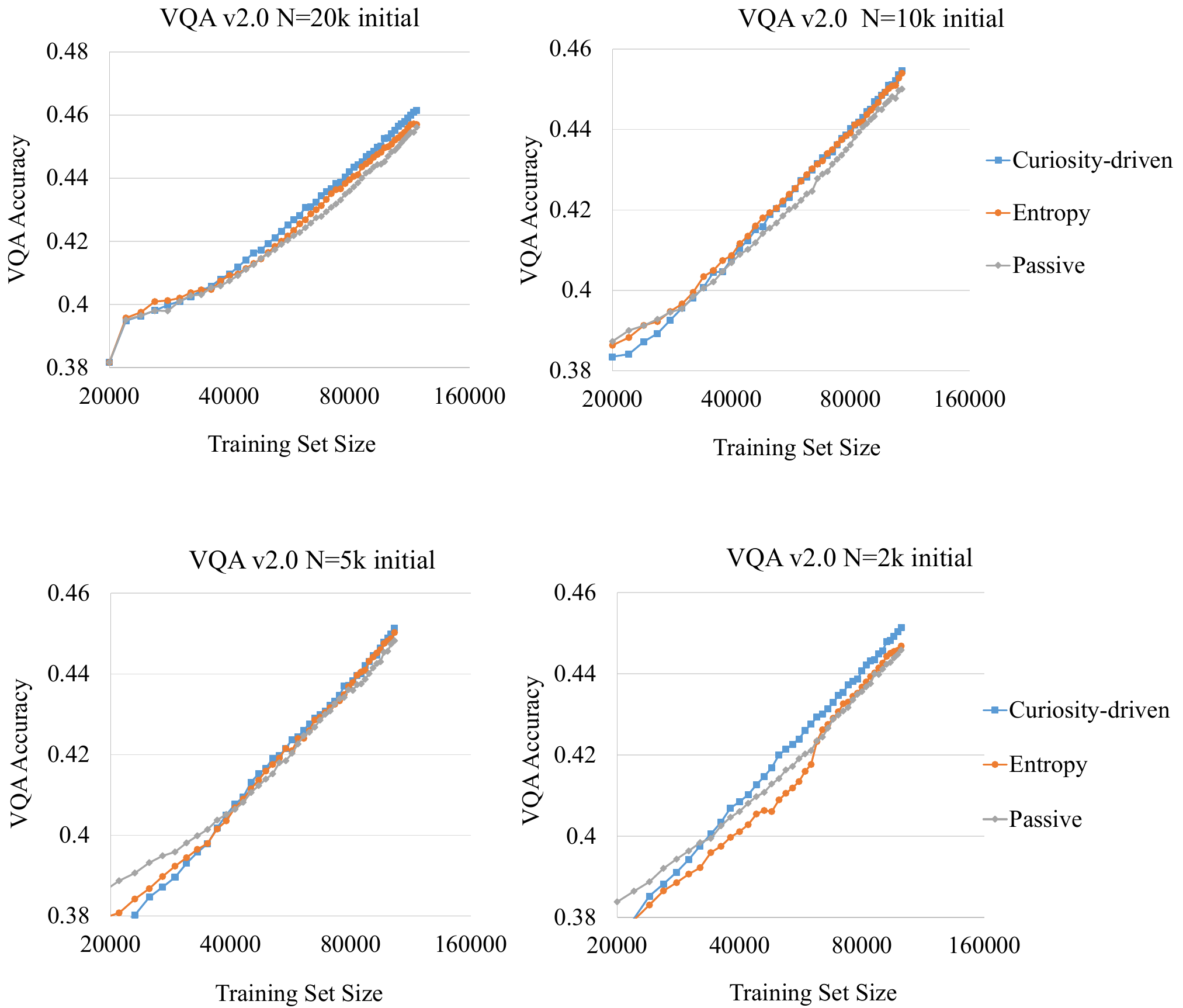}
\label{fig:al_breakpoint}
\end{figure}

On VQA v2.0, we also experiment with smaller initial training sets $N \in \{ 20\text{k}, 10\text{k}, 5\text{k}, 2\text{k}\}$ to study the impact of training set size on active learning performance. Fig.~\ref{fig:al_breakpoint} shows the results. For all initial training set sizes, the breakpoint when active learning methods start to outperform passive learning is around $30\text{k}$ to $50\text{k}$ examples. It shows that active learning methods do require a large training set size to start asking informative questions. Models with smaller initial training set sizes tend to show less and inconsistent data savings compared to $N=50\text{k}$, possibly because such models are less capable of telling informative questions from redundant ones. In addition, entropy shows fluctuating performance while curiosity-driven learning performs consistantly better than both entropy and passive learning irrespective of initial training set size.

\subsection{Goal-driven Active Learning}
\label{sec:al_goal_result}

To evaluate our goal-driven learning approach, we keep the initial training set and the pool unchanged for VQA v2.0 -- the model is allowed to ask all kinds of questions from the VQA v2.0 TRAIN split -- but will be evaluated on only ``yes/no'' questions (questions whose answers are ``yes'' or ``no'' ) in the VAL split. This task tests our proposed goal-driven active learning approach's ability to focus on achieving the goal of answering ``yes/no'' questions more accurately.

\begin{figure}[htbp]
\centering
   \caption{\textbf{Top}: Goal-driven active learning of VQA for answering only ``yes/no'' questions. Our goal-driven active learning approach outperforms passive learning and other active learning approaches. \textbf{Bottom}: Query compositions of active learning approaches, on VQA v2.0 dataset for the task of answering only ``yes/no'' questions. Our goal-driven active learning approach queries mostly ``yes/no'' questions. Best viewed in color.}
   \includegraphics[width=0.65\linewidth]{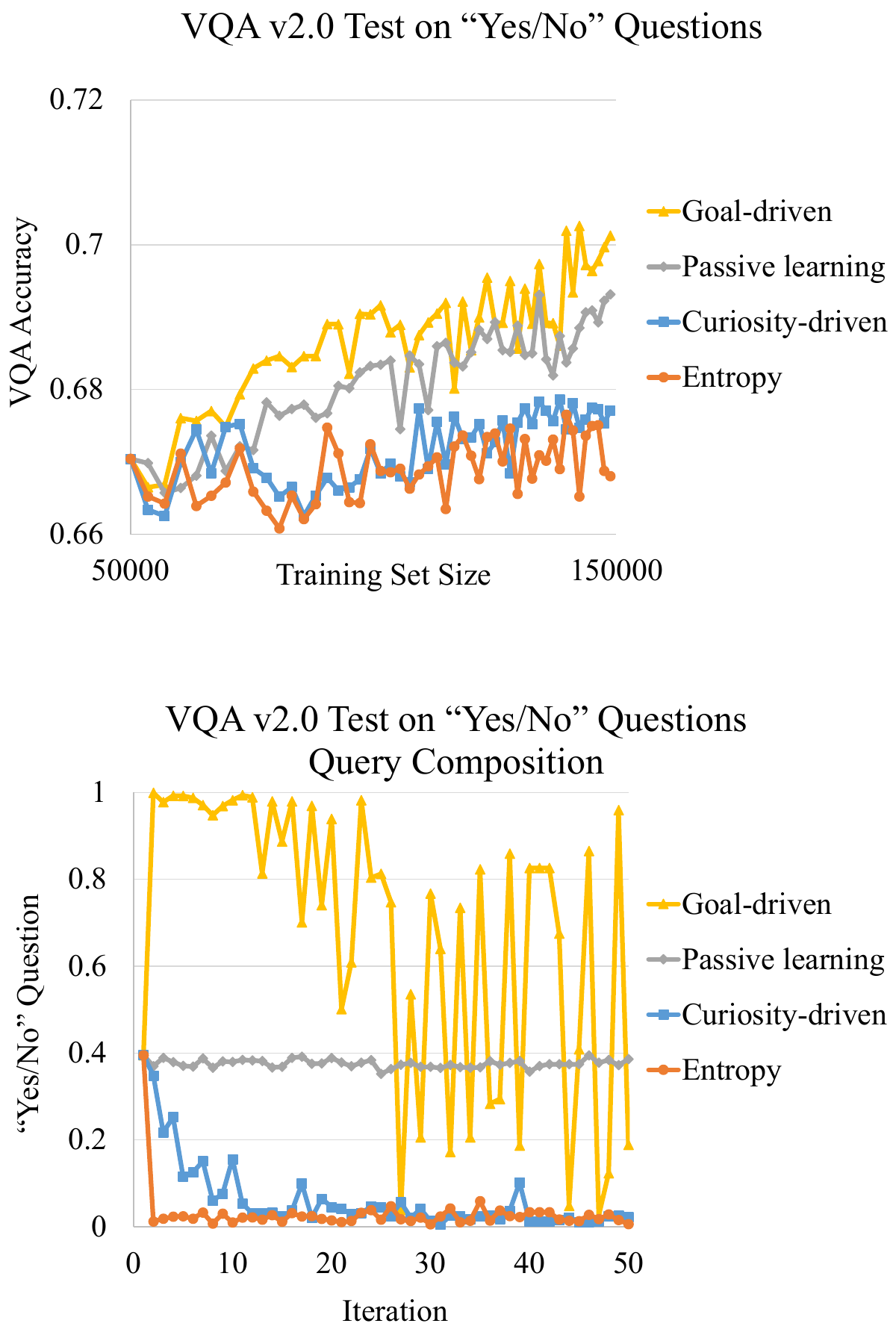}
\label{fig:al_cross_binary}
\end{figure}

Fig.~\ref{fig:al_cross_binary} (top) shows the performance of active and passive learning approaches on this task.\footnote{We also found that updating $\bm{\theta}$ from previous iteration in Algorithm~\ref{alg:al} step~\ref{step10} leads to slight overfitting that affects mutual information approximation. So for this task, $\bm{\theta}$ is initialized from scratch in every iteration.} Our goal-driven active learning approach is able to select relevant questions as queries and outperforms passive learning. Curiosity-driven and entropy approaches perform poorly. They are not aware of the task and tend to be attracted to harder, open-ended questions, which are not very relevant to the task.

Fig.~\ref{fig:al_cross_binary} (bottom) shows a closer examination of the composition of questions queried by the goal-driven learning approach compared to baseline approaches. The goal-driven learning approach queries mostly ``yes/no'' questions, which are presumably more useful for the task. Note that the approach was not told that the downstream task is to answer ``yes/no'' questions. The approach figures out which questions will be informative to ask just based on samples from the downstream task. 
It shows that the goal-driven scoring function in Eq.~\ref{eq:al_s_goal_old}, as well as the approximations in Eq.~\ref{eq:al_s_goal} are indeed effective for selecting informative questions.

% In addition, by jointly optimizing for the whole batch, instead of forming a batch by picking high scoring (Q,I) pairs independently
%Composition swings in later iterations are likely due to the effect of greedy selection based on query scoring, which could potentially be improved by using smaller batch sizes or selecting batches of queries in a holistic manner.

\begin{figure}[t]
\centering
   \caption{Goal-driven active learning of VQA for answering only ``yes/no'' questions, compared to passive learning that ``cheats'' and queries only ``yes/no'' questions. Best viewed in color.}
   \includegraphics[width=0.65\linewidth]{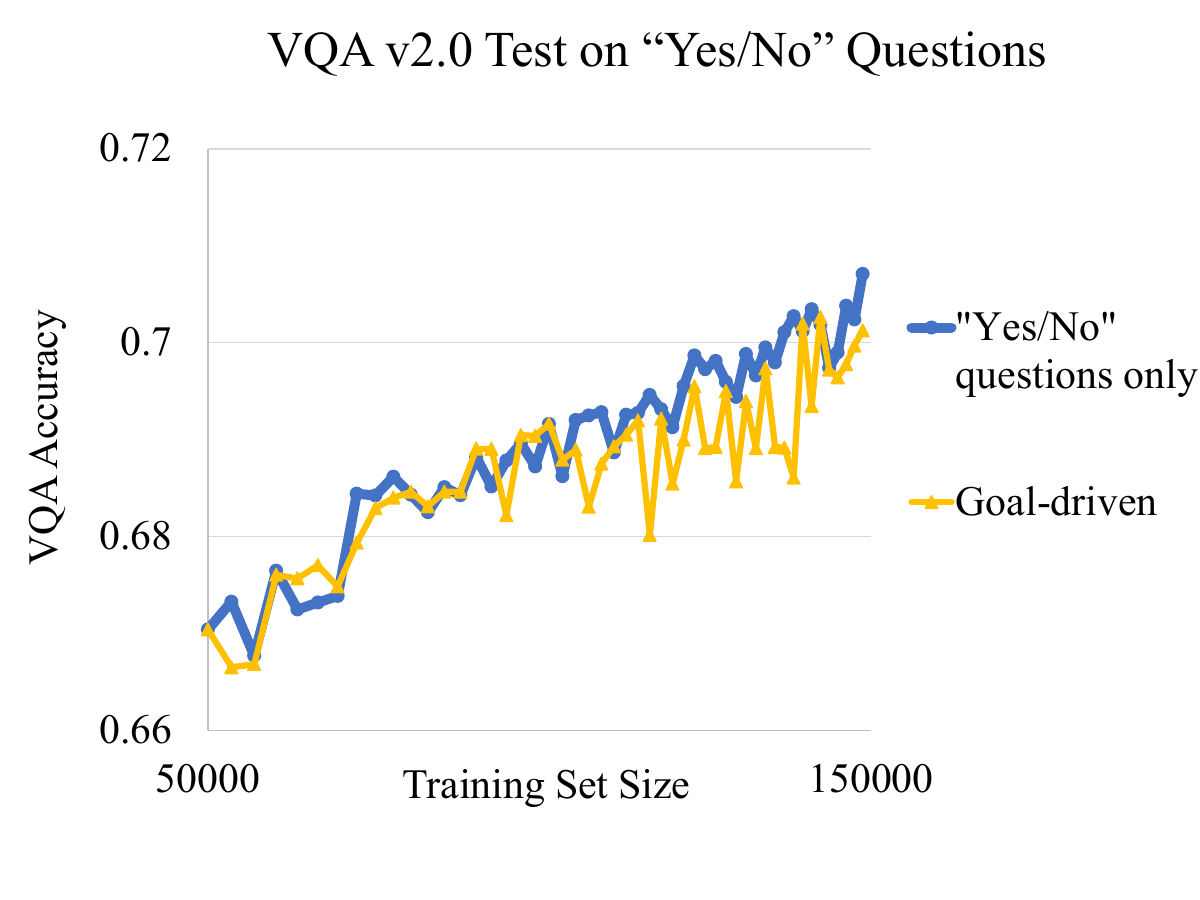}
\label{fig:al_cross_cheat}
\end{figure}

\xiao{As an ``upper bound'', it is reasonable to assume\footnote{Note that this is not necessarily the case. Even non-yes/no questions can help a VQA model get better at answering yes/no questions by learning concepts from non-yes/no questions that can later come handy for yes/no questions. For example ``Q: What is the man doing? A: Surfing'' can be as useful as ``Q: Is the man surfing? A: Yes''.} 
that ``yes/no'' questions are more desirable for this task. Imagine a passive learning method that ``cheats'': one that is aware that it will be tested only on ``yes/no'' questions, as well as knowing which questions are ``yes/no' questions in the pool, so it restricts itself to query only ``yes/no'' questions. How does our goal-driven learning approach compare with such a method that only learns from ``yes/no'' questions?  }
\xiao{Fig.~\ref{fig:al_cross_cheat} shows the results. Our goal-driven learning is able to compete with the ``cheating'' approach. 
In fact, of all 167,499 ``yes/no'' questions in the VQA v2.0 TRAIN split, goal-driven learning finds 38\% of them by iteration 25, and 50\% of them by iteration 50. That might also have made finding the remaining ``yes/no'' questions more difficult which explains the drop of the rate of ``yes/no'' question towards later iterations. We expect that a larger pool (\ie a larger VQA dataset) would reduce these issues.}

\section{Discussion}

In this work we discussed three active learning strategies -- cramming (entropy), curiosity-driven learning and goal-driven learning -- for Visual Question Answering using deep multimodal neural networks. Our results show that deep VQA models require $30\text{k}$ - $50\text{k}$ training questions for active learning before they are able to ask informative questions and achieve better scaling than randomly selecting questions for labeling. Once the training set is large enough, several active learning strategies achieve significant savings in answer annotation cost. Our proposed goal-driven query strategy in particular, shows a significant advantage on improving performance when the downstream task involves answering a specific type of VQA questions.

Jointly selecting batches of examples as queries~\cite{Sener_2017} and formulating active learning as a decision making problem~\cite{Javdani_2014} (greedily selecting the batch that reduces entropy by the most for the current iteration may not be the optimal decision) have been shown to improve optimality in active learning query selection. Combining those approaches with deep neural networks under the Bayesian Neural Network framework are promising future directions.

The pool-based active learning setup explored in this work selects unlabeled human generated question-image pairs and asks the oracle for answers. For building VQA datasets however, collecting human-generated questions paired with each image is also a substantial portion of the overall cost. 
Hence, starting from a bank of questions and an unaligned bank of images, and having the model decide which question it would like to pair with each image to use as a query would result in a further reduction in cost. Note that such a model would need to not only reason about the informativeness of a question-image pair, but also about the relevance of a question to the image~\cite{Ray_2016,Mahendru_2017}.
Evaluating such an approach would require collecting new VQA datasets with humans in the loop to give answers -- which we show would require $30\text{k}$ - $50\text{k}$ answers before the model could start selecting informative images and questions. Going one step further, we could also envision a model that generates new questions rather than selecting from a pool of questions. That would require a generative model that can perform inference to optimize for the active learning objectives. We hope that our work serves as a foundation for these future research directions.

\section*{Acknowledgements}

We thank Michael Cogswell and Qing Sun for discussions about the active learning strategies. This work was funded in part by an NSF CAREER award, ONR YIP award, Allen Distinguished Investigator award from the Paul G. Allen Family Foundation, Google Faculty Research Award, and Amazon Academic Research Award to DP. The views and conclusions contained herein are those of the authors and should not be interpreted as necessarily representing the official policies or endorsements, either expressed or implied, of the U.S. Government, or any sponsor.

\bibliography{main}
\bibliographystyle{ieee}

\pagebreak
\appendix

\section{Fast Approximation of Goal-driven Scoring Function}
\label{appendix:al_goal_approx}

In Section~\ref{sec:strategy}, we discuss our proposed goal-driven query strategy that minimizes uncertainty (entropy) on answers $A'_t$ to a given set of unlabeled test question-image pairs $(Q'_t, I'_t), t=1,2,...T$, against which the model will be evaluated. It queries $(Q,I)$ pairs which maximize:

\begin{align}
\label{eq:al_s_goal_old_appendix}
\nonumber s_{goal} &(Q,I) \\
\nonumber = & \sum_{t} \mathbb{H}(A'_t) - \mathbb{H}(A'_t|A) \\
\nonumber = & \sum_{t} \mathbb{I}(A;A'_t) \\
 = &\sum_{t} \sum_a \sum_{a'} P(A=a,A'_t=a'|Q,I,Q'_t,I'_t) \log \frac{P(A=a,A'_t=a'|Q,I,Q'_t,I'_t)}{P(A=a|Q,I)P(A'_t=a'|Q'_t,I'_t)}
\end{align}

Recall that we propose an approximation for term $P(A=a,A'_t=a'|Q,I,Q'_t,I'_t)$ as follows:

\begin{align}
\label{eq:joint_prob_approx_appendix}
\nonumber & P(A=a,A'_t=a'|Q,I,Q'_t,I'_t) \\
\nonumber & = \mathbb{E}_{\bm{\omega}} P(A=a|Q,I,\bm{\omega})P(A'_t=a'|Q'_t,I'_t,\bm{\omega}) \\
& \approx \mathbb{E}_{\bm{\omega}\sim q_\theta(\bm{\omega})} P(A=a|Q,I,\bm{\omega})P(A'_t=a'|Q'_t,I'_t,\bm{\omega})
\end{align}

Let us define four matrices $\bm{M}_1, \bm{D}_1, \bm{M}_2(t), \bm{D}_2(t)$ as follows:

\begin{align}
\bm{M}_1=
  \begin{bmatrix}
    P(A=a_1|Q,I,\bm{\omega}_1) & P(A=a_2|Q,I,\bm{\omega}_1) & \ldots & P(A=a_J|Q,I,\bm{\omega}_1)  \\
    P(A=a_1|Q,I,\bm{\omega}_2) & P(A=a_2|Q,I,\bm{\omega}_2) &  & P(A=a_J|Q,I,\bm{\omega}_2)  \\
    \vdots &   & \ddots & \vdots  \\
    P(A=a_1|Q,I,\bm{\omega}_M) & P(A=a_2|Q,I,\bm{\omega}_M) & \ldots & P(A=a_J|Q,I,\bm{\omega}_M) 
  \end{bmatrix}
\end{align}

\begin{align}
\bm{D}_1=\text{Diag} \big(
  \begin{bmatrix}
    P(A=a_1|Q,I) & P(A=a_2|Q,I) & \ldots & P(A=a_J|Q,I)  
  \end{bmatrix}
  \big)
\end{align}

\begin{align}
\bm{M}_2(t)=
  \begin{bmatrix}
    P(A_t'=a_1|Q_t',I_t',\bm{\omega}_1) & P(A_t'=a_2|Q_t',I_t',\bm{\omega}_1) & \ldots & P(A_t'=a_J|Q_t',I_t',\bm{\omega}_1)  \\
    P(A_t'=a_1|Q_t',I_t',\bm{\omega}_2) & P(A_t'=a_2|Q_t',I_t',\bm{\omega}_2) &  & P(A_t'=a_J|Q_t',I_t',\bm{\omega}_2)  \\
    \vdots &   & \ddots & \vdots  \\
    P(A_t'=a_1|Q_t',I_t',\bm{\omega}_M) & P(A_t'=a_2|Q_t',I_t',\bm{\omega}_M) & \ldots & P(A_t'=a_J|Q_t',I_t',\bm{\omega}_M) 
  \end{bmatrix}
\end{align}

\begin{align}
\bm{D}_2(t)=\text{Diag} \big(
  \begin{bmatrix}
    P(A_t'=a_1|Q_t',I_t') & P(A_t'=a_2|Q_t',I_t') & \ldots & P(A_t'=a_J|Q_t',I_t')  
  \end{bmatrix}
  \big)
\end{align}

Here $\bm{M}_1$ is an $M \times J$ matrix, $\bm{D}_1$ is a $J\times J$ matrix, $\bm{M}_2(t)$ is an $M \times J$ matrix and $\bm{D}_2(t)$ is a $J \times J$ matrix. 
With $\bm{M}_1, \bm{D}_1, \bm{M}_2(t), \bm{D}_2(t)$ we could rewrite Eq.~\ref{eq:joint_prob_approx_appendix} in matrix form:

\begin{align}
\label{eq:al_p_matrix}
\nonumber & \begin{bmatrix}
    P(A=a_1,A'_t=a_1|Q,I,Q'_t,I'_t) & \ldots & P(A=a_1,A'_t=a_J|Q,I,Q'_t,I'_t)  \\
    \vdots &  \ddots & \vdots  \\
    P(A=a_J,A'_t=a_1|Q,I,Q'_t,I'_t) & \ldots & P(A=a_J,A'_t=a_J|Q,I,Q'_t,I'_t)  
  \end{bmatrix} \\
& \approx \frac{1}{M} \bm{M}^T_1 \bm{M}_2(t)
\end{align}

Let $\text{Sum}(\cdot)$ be an operator on a matrix that sums up all elements in that matrix. Combining Eq.~\ref{eq:al_s_goal_old_appendix} and Eq.~\ref{eq:al_p_matrix}, our goal-driven scoring function can be approximately computed as follows:

\footnotesize
\begin{align}
\label{eq_al_s_goal_appendix}
\nonumber s_{goal} &(Q,I)  && \text{}\\
\nonumber = &\sum_{t} \sum_a \sum_{a'} P(A=a,A'_t=a'|Q,I,Q'_t,I'_t) \log \frac{P(A=a,A'_t=a'|Q,I,Q'_t,I'_t)}{P(A=a|Q,I)P(A'_t=a'|Q'_t,I'_t)} && \text{} \\
\nonumber \approx &\sum_{t} \text{Sum} \big\{ \big[ \frac{1}{M} \bm{M}^T_1 \bm{M}_2(t) \big] \circ \log \big[ \frac{1}{M} \bm{D_1} \bm{M}^T_1 \bm{M}_2(t) \bm{D}_2(t) \big] \big\} && \hspace{-20pt}\text{Rewriting in matrix form.} \\
\nonumber \approx & \sum_{t} \frac{1}{2} \text{Sum} \big\{ - \frac{1}{M} \bm{M}^T_1 \bm{M}_2(t) +  \frac{1}{M^2} \big[ \bm{M}^T_1 \bm{M}_2(t) \big] \circ \big[ \bm{D_1} \bm{M}^T_1 \bm{M}_2(t) \bm{D}_2(t) \big] \big\}   && \hspace{-20pt} x\log x\approx \frac{1}{2}(-x+x^2).\\
\nonumber = & \sum_{t} - \frac{1}{2} + \frac{1}{2M^2} \text{Sum} \big\{ \big[ \bm{M}^T_1 \bm{M}_2(t) \big] \circ \big[ \bm{D_1} \bm{M}^T_1 \bm{M}_2(t) \bm{D}_2(t) \big] \big\}  && \hspace{-20pt}\text{Sum of $P(A,A'_t)$ reduces to 1.} \\
\nonumber = & \sum_{t} - \frac{1}{2} + \frac{1}{2M^2} \text{Tr} \big\{\big[ \bm{M}^T_1 \bm{M}_2(t) \big] \big[ \bm{D_1} \bm{M}^T_1 \bm{M}_2(t) \bm{D}_2(t) \big]^T \big\}  && \hspace{-20pt}\text{Sum}(A\circ B)= \text{Tr} (AB^T). \\
\nonumber = & \sum_{t} - \frac{1}{2} + \frac{1}{2M^2} \text{Tr} \big[ \bm{M}^T_1 \bm{M}_2(t) \bm{D}_2(t) \bm{M}^T_2(t) \bm{M}_1 \bm{D}_1  \big] && \text{} \\
\nonumber = & \sum_{t} - \frac{1}{2} + \frac{1}{2M^2} \text{Tr} \big[ \bm{M}_1 \bm{D}_1 \bm{M}^T_1 \bm{M}_2(t) \bm{D}_2(t) \bm{M}^T_2(t)  \big]  && \hspace{-20pt}\text{Property of trace.} \\
\nonumber = & \sum_{t} - \frac{1}{2} + \frac{1}{2M^2} \text{Sum} \big\{ \big[ \bm{M}_1 \bm{D}_1 \bm{M}^T_1 \big] \circ \big[ \bm{M}_2(t) \bm{D}_2(t) \bm{M}^T_2(t) \big] \big\}  && \hspace{-20pt}\text{Tr} (AB^T)=\text{Sum}(A\circ B).  \\
\nonumber = & \sum_{t} - \frac{1}{2} + \frac{1}{2}  \mathbb{E}_{\bm{\omega}} \mathbb{E}_{\bm{\omega}'} \big[ \sum_a \frac{P(A=a|Q,I,\bm{\omega}) P(A=a|Q,I,\bm{\omega}')}{P(A=a|Q,I)}   && \hspace{-20pt}\text{Rewriting in probability form.} \\
\nonumber  & \sum_a \frac{P(A'_t=a|Q'_t,I'_t,\bm{\omega}) P(A'_t=a|Q'_t,I'_t,\bm{\omega}')}{P(A'_t=a|Q'_t,I'_t)} \big]  && \text{} \\
\nonumber = & \frac{1}{2}  \mathbb{E}_{\bm{\omega}} \mathbb{E}_{\bm{\omega}'} \big[ \sum_a \frac{P(A=a|Q,I,\bm{\omega}) P(A=a|Q,I,\bm{\omega}')}{P(A=a|Q,I)}  && \hspace{-20pt}\text{Rearranging summation. } \\
 & \sum_{t} \sum_a \frac{P(A'_t=a|Q'_t,I'_t,\bm{\omega}) P(A'_t=a|Q'_t,I'_t,\bm{\omega}')}{P(A'_t=a|Q'_t,I'_t)} \big] -\sum_{t} \frac{1}{2} && \text{} 
\end{align}
\normalsize

Which is Eq.~\ref{eq:al_s_goal} in Section~\ref{sec:strategy}. 

As stated in Section~\ref{sec:strategy}, the above equation can be computed as a dot-product between two vectors of length $M^2$. One vector is matrix $\frac{1}{M} \bm{M}_1 \bm{D}_1 \bm{M}^T_1$ expanded into a vector. It only involves pool questions $(Q,I)$. The other vector is $\frac{1}{M} \sum_{t} \bm{M}_2(t) \bm{D}_2(t) \bm{M}^T_2(t)$ expanded into a vector. It only involves test questions $(Q_t',I_t')$ and it is shared for all pool questions $(Q,I)$, so it can be precomputed for all $(Q,I)$. Precomputing $\frac{1}{M} \sum_{t} \bm{M}_2(t) \bm{D}_2(t) \bm{M}^T_2(t)$ for test questions has a time complexity of $O(TJM^2)$. Note that $\bm{D}_2(t)$ is a diagonal matrix, so multiplying $\bm{D}_2(t)$ with $\bm{M}_2^T(t)$ only takes $O(JM)$ operations. In the same way, computing $\frac{1}{M} \bm{M}_1 \bm{D}_1 \bm{M}^T_1$ for all $(Q,I)$ has a time complexity of $O(UJM^2)$. The time complexity of their dot product for all $(Q,I)$ is merely $O(UM^2)$. So the overall time complexity is $O(max(U,T)JM^2)$. The overall time complexity is linear to both dataset size $U$ and $T$ and the number of possible answers $J$, so our approach can easily scale to very large datasets and more VQA answers.

\section{Quality of Approximations}
\label{appendix:al_quality}
Our entropy, curiosity-driven and goal-driven scoring functions use 3 types of approximations

\begin{enumerate}[(a)]
\item Variational distribution $q_{\bm{\theta}}(\bm{\omega})$ as approximation to model parameter distribution $p(\bm{\omega}|\mathcal{D}_{train})$.
\item Monte Carlo sampling over $q_{\bm{\theta}}(\bm{\omega})$ for computing expectation over $p(\bm{\omega}|\mathcal{D}_{train})$.
\item Fast approximation to mutual information in Eq.~\ref{eq_al_s_goal_appendix}.
\end{enumerate}

\begin{figure}[t]
\centering
   \caption{Convergence of Monte Carlo approximation to entropy, curiosity-driven and goa-driven scoring functions in terms of rank correlation. We compute scores using Eq.~\ref{eq:al_s_entropy} (entropy), \ref{eq:al_s_curiosity} (curiosity-driven) and \ref{eq:al_s_goal_old} (goal-driven) for 200 random examples from the pool using $M \in {\{1,2,5,10,20,50,100,}$ ${200,500\}}$ samples from $q_{\bm{\theta}}(\bm{\omega})$, and compare them with $M=500$ in terms of rank correlation (Spearman's $\rho$).}
   \includegraphics[width=0.65\linewidth]{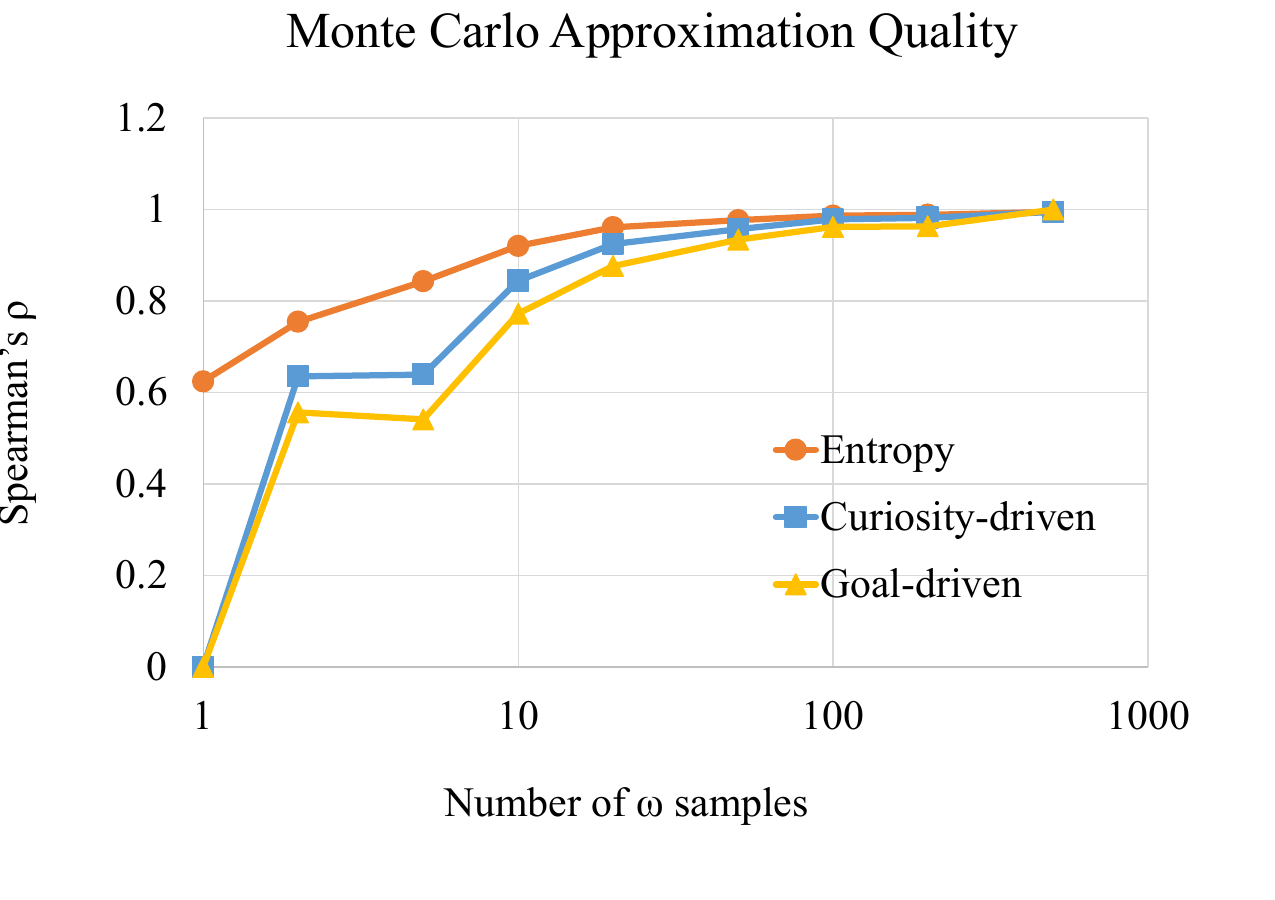}
\label{fig:al_approx_mc_corr}
\end{figure} 

For (a), since the space of model parameters is very large, it is intractable to evaluate how accurate $q_{\bm{\theta}}(\bm{\omega})$ substitutes $p(\bm{\omega}|\mathcal{D}_{train})$ for expectation computation. But nevertheless our goal-driven learning results in Section~\ref{sec:al_goal_result} suggest that Eq.~\ref{eq:al_s_goal} computed using $q_{\bm{\theta}}(\bm{\omega})$ is indeed useful for selecting relevant examples. It remains as an open problem that how to quantitatively evaluate the quality of $q_{\bm{\theta}}(\bm{\omega})$ for the purpose of uncertainty estimation and expectation computation.

\begin{figure}[t]
\centering
   \caption{Entropy, curiosity-driven and goal-driven scores of 50 examples under different numbers of model parameter samples.}
   \includegraphics[width=0.9\linewidth]{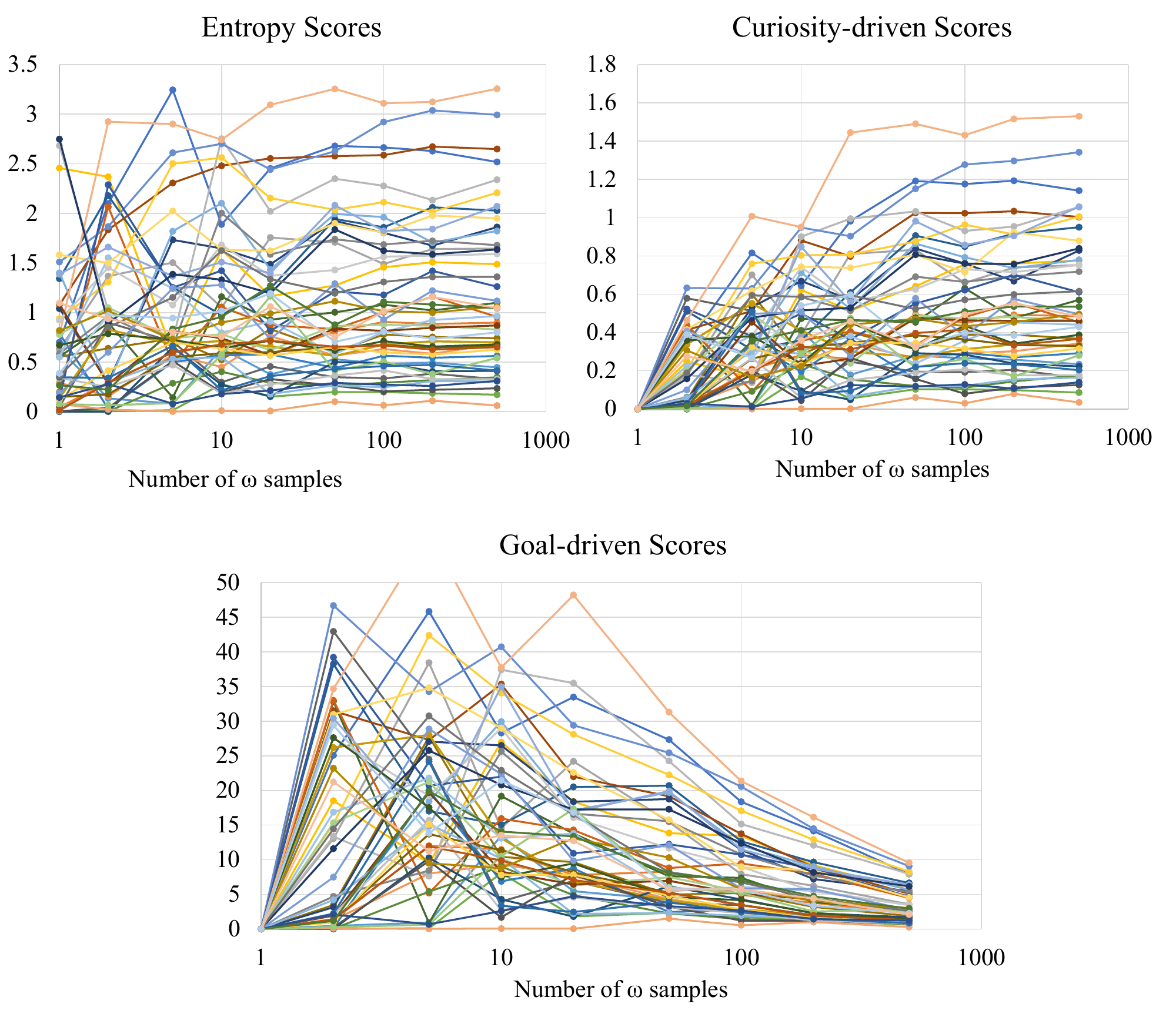}
\label{fig:al_approx_mc_score}
\end{figure}

For (b), we study the convergence patterns of Monte Carlo sampling. Specifically, given an arbitrary VQA model\footnote{For our experiments we use the model from curiosity-driven learning at iteration 50. This choice is made arbitrarily and does not change conclusions.}, we compute scores using Eq.~\ref{eq:al_s_entropy} (entropy), \ref{eq:al_s_curiosity} (curiosity-driven) and \ref{eq:al_s_goal_old} (goal-driven) for 200 random examples from the pool using $M \in \{1,2,5,10,20,50,100,$ $200,500\}$ samples from $q_{\bm{\theta}}(\bm{\omega})$, and compare them with $M=500$ in terms of rank correlation (Spearman's $\rho$). Note that we use different seeds for the different $M$ values, \ie $\omega$ samples for $M=200$ do not overlap with $\omega$ samples for $M=500$. Fig.~\ref{fig:al_approx_mc_corr} shows the results.
Entropy, curiosity-driven and goal-driven scoring functions require increasingly more samples of model parameters to converge in terms of ranking. To reach $\rho=0.9$, entropy, curiosity-driven and goal-driven scoring functions require 10, 20 and 50 samples from $q_{\bm{\theta}}(\bm{\omega})$ respectively. 
Fig.~\ref{fig:al_approx_mc_score} shows how the actual scores of examples change according to number of samples from $q_{\bm{\theta}}(\bm{\omega})$ for 50 random examples in the pool. The entropy and curiosity-driven scores seem to converge with a large number of samples. The goal-driven scores however, tend to first increase and then decrease with the number of samples and have not yet converged by $M=500$ samples, which is a limitation of the Monte Carlo sampling approach. Despite that, the relative rankings based on which the queries are selected have mostly converged. Upper- and lower-bounds of Eq.~\ref{eq:al_s_goal_old} that might improve convergence are opportunities for future research.

For (c), we plot goal-driven scores Eq.~\ref{eq:al_s_goal_old} as the x-axis versus our fast approximations Eq.~\ref{eq:al_s_goal} as the y-axis for 200 random examples from the pool using $M= \{2,5,10,20,50,100,200,$ $500\}$ samples from $q_{\bm{\theta}}(\bm{\omega})$. Because Eq.~\ref{eq:al_s_goal_old} does not scale well to large datasets, we use a subset of 200 random $(Q'_{t},I'_{t})$ pairs from the VAL split as the test domain for both Eq.~\ref{eq:al_s_goal_old} and Eq.~\ref{eq:al_s_goal}. Fig.~\ref{fig:al_approx_mc_score} shows the results. Our fast approximations are mostly linear to the goal-driven scores. The slope changes according to the number of model parameter samples $M$. That is probably because our approximation $\frac{1}{2}(-x+x^2)$ (see Section~\ref{appendix:al_goal_approx} for details) overestimates $x\log x$ for $x>1$. The rank correlations between goal-driven scores and their fast approximations remain high, \eg above $\rho>0.96$ even for $M=500$, which is sufficient for query selection.

\begin{figure}[htbp]
\centering
   \caption{Our fast approximations using Eq.~\ref{eq:al_s_goal} versus the original goal-driven scores computed using Eq.~\ref{eq:al_s_goal_old} under $M= \{2,5,10,20,50,100,200,$ $500\}$ samples of model parameters. Our approximations have high rank correlation with scores computed using the original method.}
   \includegraphics[width=0.80\linewidth]{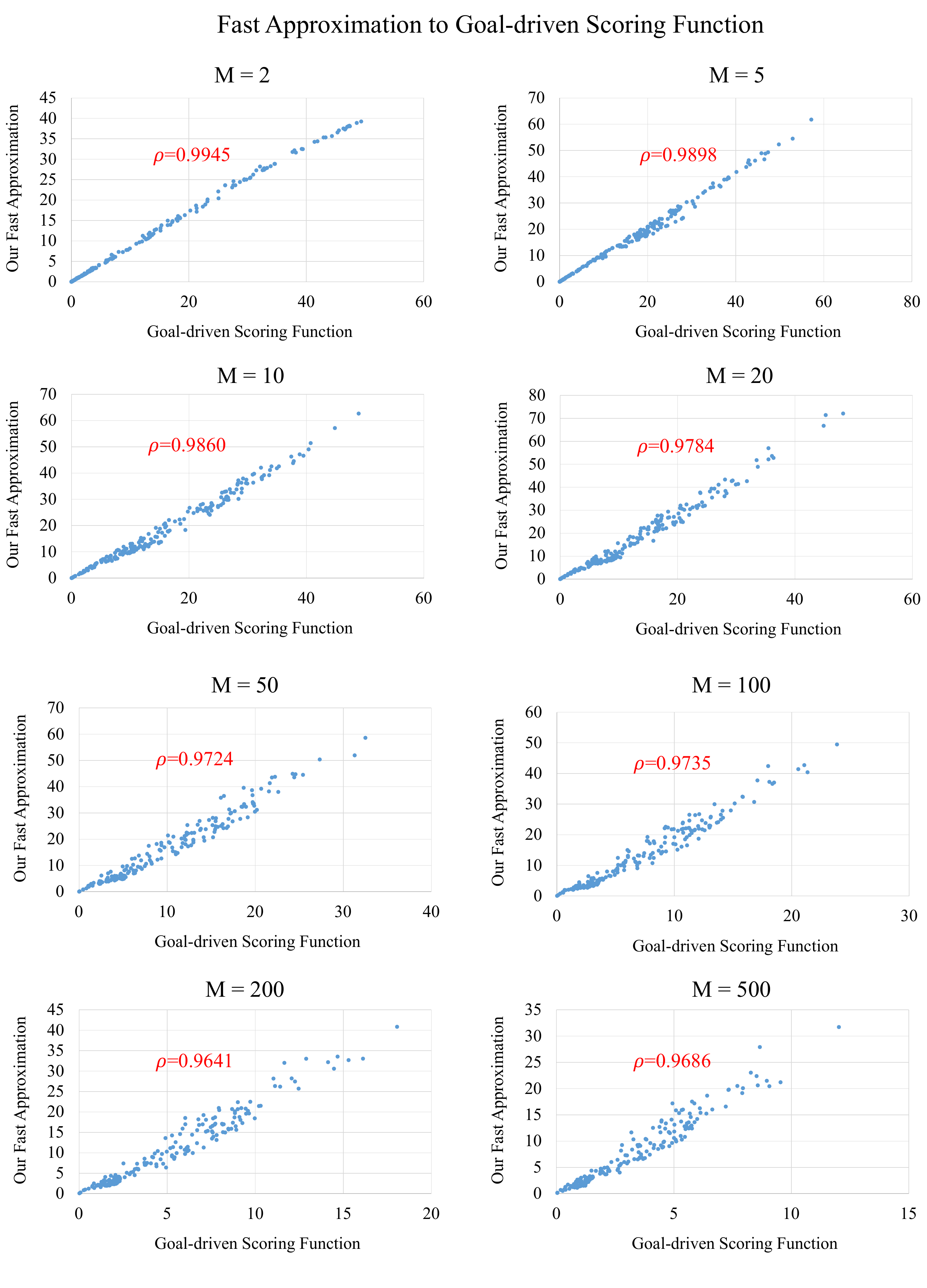}
\label{fig:al_approx_mc_score}
\end{figure}

\end{document}